\documentclass{article} %
\usepackage{iclr2024_conference,times}

\usepackage{inconsolata}
\usepackage{hyperref}
\usepackage{url}
\usepackage{booktabs}
\usepackage{graphicx}
\usepackage{times}
\usepackage{latexsym}
\usepackage{amsmath}
\usepackage{hyperref}
\usepackage{cleveref}
\usepackage{subcaption}
\usepackage{supertabular}
\usepackage{booktabs}
\usepackage{tabularx}
\usepackage{multirow}
\usepackage{multicol}
\usepackage{makecell}
\usepackage{setspace}
\usepackage{graphicx}
\usepackage{amssymb}

\usepackage{longtable}
\usepackage{tcolorbox}
\usepackage{ulem}
\usepackage{listings}

\tcbuselibrary{listingsutf8}

\newtcblisting{longlisting}[1][]{
    listing only,
    listing options={%
        basicstyle=\ttfamily\footnotesize,
        keywordstyle=\color{blue},
        stringstyle=\color{red},
        commentstyle=\color{gray},
        showspaces=false,
        showstringspaces=false,
        showtabs=false,
        frame=single,
        tabsize=4,
        breaklines=true,
        breakatwhitespace=true,
        breakautoindent=true,
        postbreak=\mbox{\textcolor{red}{$\hookrightarrow$}\space},
    },
    colback=white,
    colframe=black,
    coltitle=black,
    colbacktitle=gray!10!white,
    fonttitle=\bfseries,
    title=#1
}

\DeclareTextFontCommand{\emph}{\itshape}

\title{Sibyl: Simple yet Effective Agent Framework for Complex Real-world Reasoning}

\author{Yulong Wang$^1$\thanks{Equal Contributions.} , Tianhao Shen$^2$\footnotemark[1] , Lifeng Liu$^1$, Jian Xie$^1$  \\
$^1$ Baichuan Inc.
$^2$ College of Intelligence and Computing, Tianjin University, Tianjin, China \\
\texttt{\{wangyulong, liulifeng, richard\}@baichuan-inc.com} \\
\texttt{thshen@tju.edu.cn}
}

\iclrfinalcopy %
\begin{document}

\maketitle

\begin{abstract}

Existing agents based on large language models (LLMs) demonstrate robust problem-solving capabilities by integrating LLMs’ inherent knowledge, strong in-context learning and zero-shot capabilities, and the use of tools combined with intricately designed LLM invocation workflows by humans. However, these agents still exhibit shortcomings in long-term reasoning and under-use the potential of existing tools, leading to noticeable deficiencies in complex real-world reasoning scenarios.
To address these limitations, we introduce \emph{Sibyl}, a simple yet powerful LLM-based agent framework designed to tackle complex reasoning tasks by efficiently leveraging a minimal set of tools. Drawing inspiration from \emph{Global Workspace Theory}, \emph{Sibyl} incorporates a global workspace to enhance the management and sharing of knowledge and conversation history throughout the system. Furthermore, guided by \emph{Society of Mind Theory}, \emph{Sibyl} implements a multi-agent debate-based jury to self-refine the final answers, ensuring a comprehensive and balanced approach. This approach aims to reduce system complexity while expanding the scope of problems solvable—from matters typically resolved by humans in minutes to those requiring hours or even days, thus facilitating a shift from System-1 to System-2 thinking. \emph{Sibyl} has been designed with a focus on scalability and ease of debugging by incorporating the concept of reentrancy from functional programming from its inception, with the aim of seamless and low effort integration in other LLM applications to improve capabilities.
Our experimental results on the GAIA benchmark test set reveal that the \emph{Sibyl} agent instantiated with GPT-4 achieves state-of-the-art performance with an average score of 34.55\%, compared to other agents based on GPT-4. 
We hope that \emph{Sibyl} can inspire more reliable and reusable LLM-based agent solutions to address complex real-world reasoning tasks.\footnote{Our code is available at \url{https://github.com/Ag2S1/Sibyl-System}.}
\end{abstract}

\section{Introduction}

Large language models (LLMs) have transformed the landscape of human-computer interaction (HCI) by offering unprecedented capabilities in understanding and generating human-like text. LLM-based agents, which are systems designed to harness these models, effectively orchestrate LLM capabilities to address complex tasks \citep{xi2023rise, wang2024survey}. These agents leverage human-designed frameworks that utilize the inherent knowledge within LLMs, often employing structured workflows that maximize the potential of in-context learning and zero-shot capabilities. Such strategies allow these agents to engage in sophisticated dialogues and problem-solving scenarios that mirror human cognitive processes \citep{sumers2023cognitive}. By incorporating prior human knowledge into the workflow, LLM-based agents can process and utilize information with a level of proficiency that was previously unattainable.
  
Despite their abilities, LLM-based agents are often limited by their inability to engage in complex questions of reasoning in real-world scenarios, where the number of reasoning steps can be numerous \citep{mialon2023gaia}. While LLMs excel in simpler, quick-answer scenarios, they struggle significantly when tasks demand lengthy, complex reasoning chains, often resulting in error propagation and a steep decline in accuracy. To address these complex real-world problems, existing systems are often intricately designed, leading to complexity that makes them difficult to evolve or optimize. This complexity not only impedes their practical deployment, but also restricts their adaptability and scalability in various LLM applications, which underscores the need for an approach that has simple design while improving the reasoning capabilities of LLM-based agents.

Furthermore, long-context management also remains a significant hurdle in LLM applications. The collection of abundant external information (e.g, computation output and error messages during coding, or intricate web content) creates a high demand for managing long context sizes effectively within the LLM’s processing capabilities. However, \citet{hsieh2024ruler} found that there exists a notable gap between the claimed length (the designed maximum context size a model can handle) and the effective length (the maximum context size a model can effectively manage) that LLMs can process. On the other hand, the challenge of long contexts is further compounded by the need to integrate information from various sources and data formats, often leading to ``context dilution'' problem where valuable information is overwhelmed by the sheer volume of data \citep{xu2023retrieval, shi2024compressing}. Effective context management is crucial to ensure that LLMs can maintain focus on relevant information without being sidetracked by less pertinent details. As such, addressing this gap is not just about increasing the raw capacity of models to handle more data but also about improving their ability to discern and prioritize information that is most critical for the task at hand.

To address the identified limitations in long-term reasoning and system complexity, we introduce \emph{Sibyl}, a simple yet powerful LLM-based agent framework. The system is compartmentalized into four main modules: tool planner, external information acquisition channel, a jury based on multi-agent debate and a global workspace. Specifically, we introduce an external information acquisition channel to receive and process external information from selected tools. To efficiently compress the received information, we integrate the concept of dialogue states from task-oriented dialogue systems \citep{budzianowski2018multiwoz, quan2020risawoz, moradshahi2023x} into the channel and design a representation language for the compressed information. This adaptation allows for the selective compression of external information, focusing only on incremental details pertinent to solving problems, which diverges from the traditional method of simply appending external information to the conversational history. By doing so, it not only elevates the quality and relevance of the information processed by LLMs but also conserves context length, allowing for more extended steps in reasoning. In addition, we design a global workspace inspired by the \emph{Global Workspace Theory} \citep{baars1993cognitive, baars2005global} that facilitates seamless information sharing among the modules, and a multi-agent debate-based jury under the guidance of \emph{Society of Mind Theory} \citep{minsky1988society} that encourages self-refinement before the final response.

The inner design of \emph{Sibyl} is inspired by functional programming principles, emphasizing reusability and statelessness between operations. This is realized through the use of QA functions instead of dialogues in internal LLM inference requests, allowing for each LLM inference to operate independently without the need to maintain a persistent state. By reducing inner dependency on LLM requests and maintaining a simple structure, we hope that \emph{Sibyl} can be easily reused to facilitate and inspire other LLM-based applications to improve their reasoning capabilities and achieve the shift from System-1 (rapid and intuitive) to System-2 (slow and delibrate) thinking.

We evaluated the \emph{Sibyl} agent instantiated by GPT-4o API (text only) in the GAIA benchmark test set, which is carefully designed to probe the depth and robustness of reasoning through a diverse set of real-world questions \citep{mialon2023gaia}. \emph{Sibyl} agent achieves an impressive average score of 34.55\% on the GAIA test set, outperforming the previous state-of-the-art method based on AutoGen \citep{wu2023autogen} and demonstrating its superior reasoning capabilities. In contrast, widely used systems like AutoGPT-4 \citep{gravitasauto} achieve only an 5\% average score. Notably, \emph{Sibyl} agent secures scores of 32.7\% and 16.33\% in the more challenging level 2 and level 3 scenarios of GAIA, respectively. These results represent significant relative improvements of 13\% and 12\% over the prior state-of-the-art method, underscoring \emph{Sibyl}’s proficiency in managing intricate, long-term reasoning tasks.

Our contributions are as follows:
\begin{itemize}
    \item We propose \emph{Sibyl}, a simple and powerful LLM-based agent framework that embodies a design philosophy centered on simplicity, modularity, and reusability by promoting stateless interactions during inference time, which facilitates ease of debugging and enhancement.
    \item We develop an external information acquisition channel accompanied by a representation language specifically tailored to efficiently gather and selectively compress external information. Drawing inspiration from \emph{Global Workspace Theory} and \emph{Society of Mind Theory}, we also introduce a global workspace that facilitates effective information sharing across modules, and a multi-agent debate-based jury that promotes self-reflection.
    \item The experimental results on the GAIA benchmark test set demonstrate that the \emph{Sibyl} agent achieves new state-of-the-art performance, particularly in the challenging Level 2 and Level 3 scenarios, which underscores the improvement of \emph{Sibyl} in solving complex reasoning tasks.
\end{itemize}

\begin{figure}[t]
    \centering
    \includegraphics[width=\linewidth]{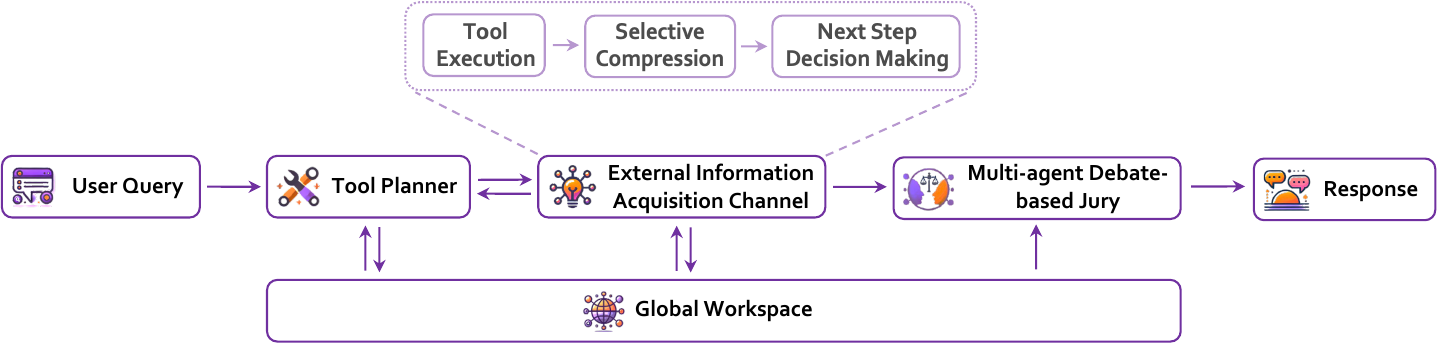}
    \caption{The overall pipeline of \emph{Sibyl} framework.}
    \label{fig:sibyl_architecture}
\end{figure}

\section{The Sibyl Framework}
In this section, we provide a overview of the \emph{Sibyl} framework, focusing on its design philosophy and fundamental modules. As shown in Figure \ref{fig:sibyl_architecture}, given a user query, \emph{Sibyl} starts with the tool planner which aims to select appropriate tools, functions, and parameters tailored to each specific subtask. Then, we design an external information acquisition channel to call the tools and selectively compress the external information returned by tool execution. Inspired by \emph{Society of Mind Theory}, we incorporate a multi-agent debate-based jury to achieve self-correction and a global workspace to seamlessly share information and enable effective collaboration across all modules. The prompts for each module are described in Appendix \ref{sec:prompts}.

\subsection{Design Philosophy}
The \emph{Sibyl} framework is constructed around a core design philosophy that focuses on reducing complexity while enhancing the functional capabilities of LLM-based agents. This philosophy is implemented through several strategic approaches to restructure LLM operations.

\paragraph{Human-oriented Browser Interface Instead of Retrieval Augmented Generation}
In conventional LLM-based agent setups, Retrieval Augmented Generation (RAG) often leads to significant information loss due to the limitation of retrieval process, which can lose the sequential information and connection of chunks in long text and have to trade off between information precision and recall. Retrieving information at a coarse level yields a wider range of data but with less precision, while a fine-level retrieval approach ensures a more detailed dataset, albeit sacrificing speed. Inspired by the success of WebGPT \citep{nakano2021webgpt}, \emph{Sibyl} addresses this by adopting a human-oriented browser interface, shifting away from RAG's constraints towards a more intuitive, human-like method of information retrieval. This form of information gathering is crucial as it retains the relational dynamics of the text, preserving more context and depth in the data accessed by the agent.

\paragraph{Question Answering Function Instead of Dialogues}
Recent agent frameworks such as AutoGen \citep{wu2023autogen} utilize dialogue as the primary mode of communication between different modules. This design is intended to mimic human conversational patterns, making the interaction more natural and user-friendly. However, dialogues are stateful and can create complex dependencies among various LLM inference calls within a session, complicating the debugging and prompt engineering processes significantly. \emph{Sibyl} replaces this with a stateless, reentrant QA function that decouples individual LLM inference requests. This transformation significantly simplifies the architecture of the system, facilitating easier maintenance and modification while allowing each component to operate independently without legacy constraints from previous interactions.

\paragraph{Less Universal Tools Instead of More Specialized Tools}
\emph{Sibyl} centralizes its functionalities around two primary tools: the Web browser and Python environments. It aligns the browser’s interface more closely with human interaction modes, such as using page navigation commands (page down/up), click, and search shortcuts (ctrl+f, ctrl+g). This approach moves away from reliance solely on web crawlers and full-page content parsing, aiming for a more selective and relevant data interaction method that mimics human web usage patterns.

\paragraph{From System-1 to System-2 Thinking}
The \emph{Sibyl} framework places a strong emphasis on enhancing capabilities for long-term memory, planning, and error correction--elements vital for complex, long-distance reasoning.

\begin{itemize}
    \item \textbf{Shared Long-term Memory as First-class Citizen:} \emph{Sibyl} incorporates a global workspace that all modules share, which is designed from the ground up and stores information with an incremental state-based representation language. This language selectively compresses past events, adding only information increments relevant to problem solving, rather than simply appending all incoming data.
    \item \textbf{Planning and Self-correction:} \emph{Sibyl} summarizes the outcomes from its tools, and plans subsequent steps based on the assessment of current progress. This involves strategic thinking about which pieces of information are necessary and how they should be processed moving forward. In addition, \emph{Sibyl} introduces a ``Jury'' mechanism, which utilizes a multi-agent debate format for self-critique and correction. This process allows the model to utilize the information stored in the global workspace efficiently to refine responses and ensure accurate problem solving.
\end{itemize}

Through these principles, \emph{Sibyl} aims to advance the development of LLM-based agents, shifting towards a model that is not only more aligned with human cognitive processes but also more adaptable and capable of handling the complexities of real-world applications. This holistic approach is key to achieving a deeper, more nuanced interaction with information, leading to better-informed, more reliable decision-making for LLM-based agents.

\subsection{Tool Planner}
The tool planner in \emph{Sibyl} is a specialized module designed to select the most suitable (tool, function, parameters) triplets and parameters to address incoming queries step by step. Given the planning prompt, it assesses the given query and any associated step history to determine the most effective tools, functions, and parameters for execution. If the query can be resolved straightforwardly and does not require additional tools, the planner can also select ``None'', indicating that no additional tools are required for that particular step.

\subsection{External Information Aquisition Channel}

The external information acquisition channel plays a pivotal role in enhancing the agent's ability to process and utilize information effectively. It first obtains the output of the tool planner, which specifies the tool, function, and parameters to be used. Upon receiving these directives, the appropriate tools are activated to gather and return the needed information.

Then, the channel performs analysis to extract and verify relevant information for the query. This involves:

\begin{itemize}
    \item Analyzing the tool results to extract relevant data that directly contributes to answering the question.
    \item Verifying the extracted information against the original question to ensure its accuracy and relevance.
    \item Recording new facts only if they provide unique and necessary information that has not already been captured in the step history.
    \item Deciding if the tool result sufficiently answers the query or if further data gathering is necessary.
\end{itemize}

If the information from the current step is insufficient, the channel plans a follow-up step to gather more data, choosing the next tool and query that will efficiently lead to the ultimate goal. This planning is detailed, focusing on minimizing unnecessary steps and emphasizing direct and efficient methods to gather required information.

Considering that the information obtained is usually lengthy and noisy, we propose a representation language to selectively compress the information inspired by \citet{lee2024benchmark}. Unlike simple data appending, selective compression involves integrating only those pieces of information that directly contribute to resolving the query at hand. This is exactly in line with the goal of the dialogue state tracking module in task-oriented dialogue systems, which is designed to selectively compress past information related to accomplish tasks. This approach not only minimizes data redundancy but also enhances the relevance and quality of information maintained in the system's memory. Specifically, we design a structured output in this step that includes sections for recording the incremental factual information compared to history, explanations of why choosing the next step and how it contributes to answering the question, and the detailed plan of the next step.

\subsection{Multi-agent Debate-based Jury}
The \emph{Society of Mind Theory}, proposed by Marvin Minsky, provides a fundamental underpinning for the design of the jury system in the \emph{Sibyl} framework. It posits that the mind is composed of a multitude of semi-autonomous agents, each responsible for different aspects of intellectual operation, making complex cognitive processes emerge from the interactions and negotiations between simpler, specialized processes. These agents work in concert, albeit through a decentralized process of negotiation and cooperation, much like a society \citep{minsky1988society}. This inspired the multi-agent debate-based jury mechanism in the \emph{Sibyl} framework, where multiple agents can discuss and analyze problems, mimicking the cooperative yet independent interaction of Minsky’s mental agents.

Here we instantiate the jury with a minimal implementation using AutoGen \citep{wu2023autogen}, where two primary roles are defined:
\begin{itemize}
    \item \textbf{Actor}: This agent attempts to answer the question, explain their thought process in detail, and consider feedback from others critically.
    \item \textbf{Critic}: The role of this agent is to identify logical or intellectual errors in the actor’s reasoning.
\end{itemize}

These roles allow for a structured yet flexible interaction and play a vital role in ensuring the logical coherence and intellectual integrity of the responses provided. We leave the exploration of diverse organizational forms for LLM-based agents for future work, including collaborative models like advisory councils.

In addition, the \emph{Sibyl} framework employs a majority vote ensemble method to enhance the stability and quality of the output answers. This method aggregates decisions or suggestions from multiple inferences, using their consensus to finalize the answer. This approach is particularly effective in mitigating individual errors in agent responses, leading to more reliable and accurate problem solving.

\subsection{Global Workspace}
The concept of the global workspace in \emph{Sibyl} is deeply influenced by \emph{Global Workspace Theory} \citep{baars1993cognitive, baars2005global}, which suggests that the brain consists of many specialized processes or modules that operate simultaneously, with a significant portion functioning unconsciously. Attention serves as a spotlight that elevates some of these unconscious activities to conscious awareness within the global workspace. This workspace acts as a crucial hub for the broadcasting and integration of information, allowing its distribution across various modules. This theoretical backdrop inspires the design of the global workspace as an integrative platform for various modules within the \emph{Sibyl} framework, facilitating seamless information sharing and a comprehensive understanding of complex problems.

The global workspace acts as a central hub where different modules can broadcast their outputs and insights. This mechanism ensures that despite the modular nature of the system, there is a cohesive and unified approach to problem solving. In addition, information within the global workspace is well-structured and denoised, which not only ensures that the data is easy to access and manipulate by LLMs but also simplifies the debugging and case-analysis processes for human developers. By implementing this global workspace, the framework supports complex information handling and long reasoning sequences, facilitating the evolution from rapid, reflexive responses (System-1 thinking) to more deliberate and structured problem solving (System-2 thinking).

\section{Experiments}
\paragraph{Datasets}
We conducted our evaluations on the GAIA dataset \citep{mialon2023gaia}, a benchmark tailored for general AI assistants. The GAIA dataset is designed to reflect tasks that align closely with human perceptual capabilities, including visual, auditory, and textual modalities. This dataset challenges general-purpose assistants with complex reasoning tasks that even require dozens of steps to solve, similar to those encountered by humans. Such a design significantly magnifies the impact of error propagation, providing a robust evaluation of problem-solving and error-handling capabilities for LLM-based agents.

\subsection{Settings}
In our experimental setup for the \emph{Sibyl} framework, we utilized two primary tools: a web browser and a Python-based code interpreter. Details of these tools are described in the Appendix \ref{sec:appendix_tools}. To balance budget constraints and time cost, we only use the GPT-4o API within the text modal, and limit the maximum number of reasoning steps of the model to 20.

\subsection{Baselines}
We compared it against several established baselines on GAIA: 1) GPT-4 \citep{achiam2023gpt} with and without manually configured plugins, 2) AutoGPT-4 \citep{gravitasauto}, which integrates AutoGPT with a GPT-4 backend, 3) An LLM-based agent implemented by AutoGen \citep{wu2023autogen}, a framework designed for automating complex multi-agent scenarios, and 4) FRIDAY \citep{wu2024copilot}, an advanced agent utilizing OS-Copilot for automating a broad spectrum of computer tasks, capable of interfacing with various operating system elements including web browsers, code terminals, and multimedia.

\subsection{Main Results}

\begin{table}[h]
\centering
\begin{minipage}{0.45\linewidth}
\centering
\resizebox{\linewidth}{!}{
\begin{tabular}{@{}lcccc@{}}
\toprule
\textbf{Model} & \textbf{Level 1} & \textbf{Level 2} & \textbf{Level 3} & \textbf{Overall} \\ 
\midrule
AutoGen         & \textbf{47.31}            & 28.93            & 14.58            & 32.33            \\ 
FRIDAY          & 40.86            & 20.13            & 6.12             & 24.25            \\ 
AutoGPT4        & 15.05            & 0.63             & 0.00                & 5.00                \\ 
GPT4 Turbo      & 9.68             & 6.92             & 0.00                & 6.67             \\ 
GPT4 w/ plugins  & 30.30             & 9.70              & 0.00                & 14.6             \\ 
GPT4            & 9.68             & 1.89             & 0.00                & 4.00                \\ 
GPT3.5          & 4.30              & 1.89             & 2.08             & 2.67             \\ 
\midrule
\emph{Sibyl}           & \textbf{47.31}            & \textbf{32.70}             & \textbf{16.33}            & \textbf{34.55}            \\ 
\bottomrule
\end{tabular}}
\caption{Performance on the GAIA test set.}
\label{tab:test_results}
\end{minipage}
\hspace{0.05\linewidth}
\begin{minipage}{0.45\linewidth}
\centering
\resizebox{\linewidth}{!}{
\begin{tabular}{@{}lcccc@{}}
\toprule
\textbf{Model} & \textbf{Level 1} & \textbf{Level 2} & \textbf{Level 3} & \textbf{Overall} \\ 
\midrule
AutoGen         & 54.72            & \textbf{38.37}            & \textbf{11.54}            & 39.39            \\ 
FRIDAY          & 45.28            & 34.88            & \textbf{11.54}            & 34.55            \\ 
AutoGPT4        & 13.21            & 0.00                & 3.85             & 4.85             \\ 
GPT4 Turbo      & 20.75            & 5.81             & 0.00                & 9.70             \\ 
GPT4 w/ plugins  & 30.30             & 9.70              & 0.00                & 14.60             \\ 
GPT4            & 15.09            & 2.33             & 0.00                & 6.06             \\ 
GPT3.5          & 7.55             & 4.65             & 0.00                & 4.85             \\ 
\midrule
\emph{Sibyl}           & \textbf{60.38}            & 36.05            & \textbf{11.54}           & \textbf{40.00} \\
\bottomrule
\end{tabular}}
\caption{Performance on the GAIA validation set.}
\label{tab:validation_results}
\end{minipage}
\end{table}

\begin{figure}[h]
    \centering
    \begin{minipage}{.55\linewidth}
    \centering
    \resizebox{\linewidth}{!}{
    \begin{tabular}{c|ccc}
    \toprule
    \textbf{Level} & \textbf{Correct?} & \textbf{\makecell{Avg. Steps of \\ Human}} & \textbf{\makecell{Avg. Steps of \\ \emph{Sibyl} Agent}} \\
    \midrule
    \multirow{2}{*}{\textbf{1}} & $\times$ & 6.43 & 5.86 \\
     & $\checkmark$ & 4.72 & \textbf{2.69} \\
    \midrule
    \multirow{2}{*}{\textbf{2}} & $\times$ & 7.93 & 11.73 \\
     & $\checkmark$ & 8.06 & \textbf{6.03} \\
    \midrule
    \multirow{2}{*}{\textbf{3}} & $\times$ & 12.52 & 13.52 \\
     & $\checkmark$ & 16.67 & \textbf{6.33} \\
    \midrule
    \multirow{2}{*}{\textbf{Overall}} & $\times$ & 8.68 & 10.90 \\
     & $\checkmark$ & 6.83 & \textbf{4.42} \\
    \bottomrule
    \end{tabular}
    }
    \captionof{table}{Average steps needed by human and \emph{Sibyl} agent on the GAIA validation set.}
    \label{table:comparison}
    \end{minipage}%
    \hspace{0.05\linewidth}
    \begin{minipage}{.35\linewidth}
        \centering
        \includegraphics[width=\linewidth]{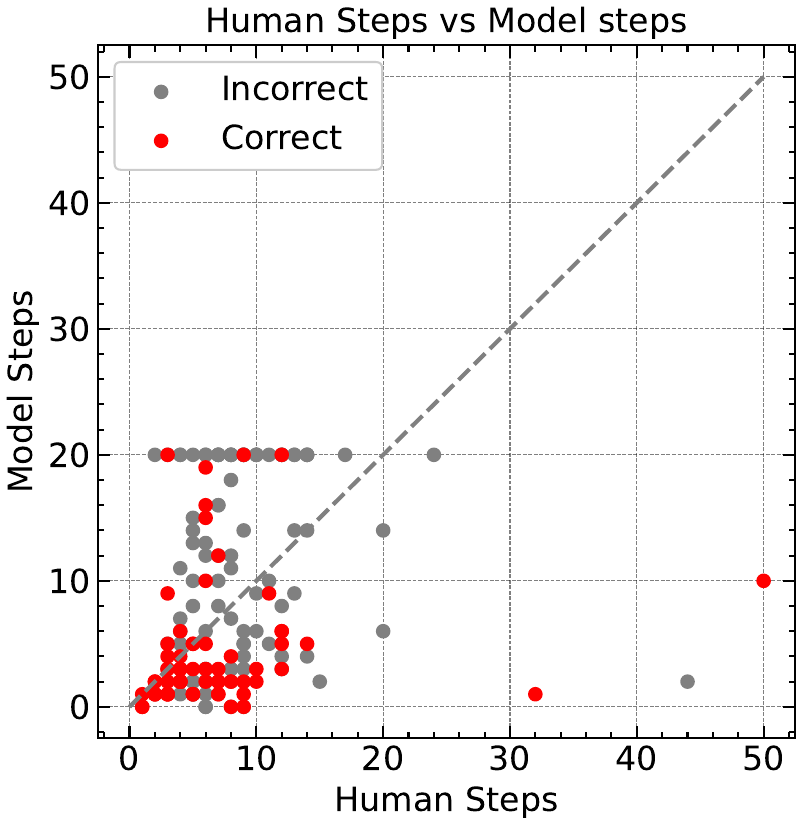}
        \caption{Steps used to solve questions for human and \emph{Sibyl} agent on the GAIA validation set.}
        \label{fig:accuracy_by_steps}
    \end{minipage}
\end{figure}

We present the results of our experiments on both the test and validation sets of the GAIA dataset in table \ref{tab:test_results} and \ref{tab:validation_results}, respectively. Our findings highlight that \emph{Sibyl} outperforms other models in the GAIA private test set, particularly in the more challenging Level 2 and Level 3 scenarios. This improvement is noteworthy given that these levels require longer steps for humans to solve, demonstrating \emph{Sibyl}'s enhanced ability to mitigate error propagation in complex reasoning processes. Furthermore, the comparison of performances on the test and validation sets indicates that \emph{Sibyl} exhibits superior generalization capabilities. The smaller decline in accuracy from validation to test set (40.00 to 34.55) compared to AutoGen (39.39 to 32.33) and FRIDAY (34.55 to 24.25) underscores \emph{Sibyl}'s robustness across different operational environments.

We also analyze the number of steps required to solve problems compared to humans. As shown in Table \ref{table:comparison}, in problems where the \emph{Sibyl} agent was correct, it consistently outperformed humans in efficiency, using significantly fewer steps in levels 1 through 3. This underscores the \emph{Sibyl}'s capability to streamline decision-making processes effectively. In addition, further insights are gained from Figure \ref{fig:accuracy_by_steps}, which plots the number of steps taken by humans versus those taken by the \emph{Sibyl} agent for each question in the GAIA validation set. These results demonstrate that despite being limited to a maximum of 20 reasoning steps, the \emph{Sibyl} agent exhibited a high level of reasoning efficiency, indicating strong capability to mitigate unnecessary reasoning and suppress error propagation.

\subsection{Ablation Studies}

\begin{table}[t]
\centering
\begin{tabular}{@{}lcccc@{}}
\toprule
\textbf{Configuration} & \textbf{Level 1} & \textbf{Level 2} & \textbf{Level 3} & \textbf{Overall} \\ 
\midrule
\emph{Sibyl}                  & 60.38            & 36.05            & 11.54            & 40.00            \\ 
w/o multi-agent debate-based jury & 49.06            & 37.21            & 11.54            & 36.97            \\ 
w/o majority vote-based ensemble           & 58.49            & 31.39            & 10.26            & 36.77            \\ 
\bottomrule
\end{tabular}
\caption{Results of ablation studies on the GAIA validation set.}
\label{tab:ablation_results}
\end{table}

We further conduct the ablation studies on the validation set of GAIA to investigate the individual contributions of specific components within the \emph{Sibyl} framework. We mainly focus on two main components: the multi-agent debate-based jury and the majority vote-based ensemble. As shown in \ref{tab:ablation_results}, the removal of the multi-agent debate component resulted in a notable performance drop in the Level 1, which indicates that this component significantly boosts the accuracy for basic question types. However, its removal did not markedly affect the performance in the more challenging Levels 2 and 3. This suggests that while the multi-agent debate is crucial for resolving simpler queries effectively, it does not substantially influence the outcome of more complex reasoning tasks.

To evaluate the impact of the ensemble component, we report the average accuracy across three separate runs to ensure a fair comparison against the ensemble configuration. The individual runs yielded overall accuracies of 40.61, 34.55, and 35.15, respectively, with an average of 36.77, compared to the ensemble's overall performance of 40.00. While one run did slightly exceed the ensemble result, the ensemble generally provided more stable and consistent outcomes. This demonstrates that although the ensemble does not guarantee superior performance in every instance, it effectively enhances result stability and reliability across varied runs.

\subsection{Discussion}

In this section, we will explore some insights gained from the development of \emph{Sibyl}. We hope that these insights can inspire future LLM-based agent work with more powerful reasoning capabilities.

\paragraph{Challenges in Complex Reasoning} Complex reasoning in real-world applications is inherently challenging due to the high risk of error propagation. Even with an 80\% accuracy rate at each step, the probability of maintaining this accuracy consistently across 20 steps plummets to merely around 1\%. This exponential increase in error risk highlights the critical nature of maintaining high accuracy at every step in reasoning. Furthermore, excessive errors can cause a series of retries which consume significant portions of the context, submerge useful information and hamper the problem-solving process.

\paragraph{Strategic Approaches to Mitigating Errors}
The decomposition of complex reasoning into simpler, manageable substeps is vital, as improving the success rate of each step can significantly mitigate error propagation. Selective compression of historical information plays a key role here, as most of the data accumulated during web navigation and previous interactions do not directly contribute to solving the problem at hand. This approach not only streamlines information processing, but also focuses on maintaining only the most pertinent data, enhancing overall system efficiency.

\paragraph{Importance of Debug-oriented Design}
A robust debug-oriented design is essential for reducing debugging costs and facilitating rapid system iteration. By limiting the introduction of state and striving for decoupling between components and even individual LLM inference requests, the system's maintainability and adaptability are significantly improved. Given that \emph{Sibyl} follows the combinator pattern, it can be seamlessly integrated as a low-cost enhancement into existing frameworks, easily replacing the vanilla GPT-4 API. This design can make it more flexible in various LLM applications.

\paragraph{Optimizing Tool Usage}
Considering the aforementioned error propagation in complex reasoning, optimizing existing tools is often more crucial than adding new ones. The potential of tools such as web browsers is far from fully realized; current LLM agents do not yet match human capabilities in terms of content visibility and operational scope on web platforms. Enhancing these tools to fully exploit their capabilities can provide substantial improvements in performance and utility, thus driving forward the sophistication and effectiveness of AI systems in complex environments.

\section{Related Work}
The integration of LLMs into autonomous agents marks a significant advancement in the field of artificial intelligence. These agents, capable of sensing their environment, making decisions and taking actions, are at the forefront of pushing AI towards Artificial General Intelligence (AGI) \citep{xi2023rise, wang2024survey}. These agents, often referred to as LLM-based agents, are increasingly prevalent across a variety of domains, demonstrating the potential of LLM applications in complex scenarios.

Most LLM-based agents are designed for specific applications, highlighting their adaptability but also suffering from a potential limitation in versatility. These applications include mathmatical problem solving \citep{gou2023tora, swan2023math}, coding \citep{yang2024swe, zheng2024opencodeinterpreter}, role-playing \citep{shao2023character, shen2023roleeval}, and social simulation \citep{park2023generative, gao2023s}. To take a step further towards general purpose LLM-based agents that are capable of various general tasks, open-source communities have developed some LLM-based agent framework, such as Langchain \citep{Chase-LangChain-2022}, BabyAGI \citep{nakajima2023babyagi} and AutoGPT \citep{gravitasauto}. Equipped with tools and structured frameworks, these agents can competently handle relatively straightforward tasks with human-like capabilities. However, their proficiency in tackling complex real-world challenges remains comparatively limited. This gap indicates the need for further enhancements in general-purpose LLM-based agents to address more intricate problems effectively.

\section{Conclusion}
We introduce \emph{Sibyl}, an agent framework designed to enhance the capabilities of LLMs in complex reasoning tasks. Through the integration of modular design and a global workspace for information sharing and collaboration, \emph{Sibyl} aims to facilitate the transition of LLM-based agents from rapid and intuitive System-1 thinking to slow and delibrate System-2 thinking. Our experimental results on the GAIA benchmark show that the \emph{Sibyl} agent instantiated by GPT-4 outperforms existing state-of-the-art solutions, demonstrating the effectiveness of our proposed framework. We hope that \emph{Sibyl} can contribute to the promotion of LLM applications to have better capabilities in handling complex real-world tasks.

\bibliography{iclr2024_conference}
\bibliographystyle{iclr2024_conference}

\newpage
\appendix
\section{Details of Tools}
\label{sec:appendix_tools}
In the development of \emph{Sibyl}, we reuse the tools from the AutoGen \citep{wu2023autogen}. Below is a detailed description of each tool and its functions, providing insights into how these tools are utilized within the framework to process and interact with web content efficiently.

\paragraph{Web Browser}
The web browser tool in \emph{Sibyl} is designed to perform a variety of functions that facilitate interaction with web pages. These functions are shown in table \ref{tab:web_browser_functions}.

\begin{table}[h]
\centering
\begin{tabular}{@{}ll@{}}
\toprule
\textbf{Function} & \textbf{Description} \\ 
\midrule
informational\_web\_search & Search for information based on a query and return results. \\ 
navigational\_web\_search & Navigate directly to a specific URL based on a search query. \\ 
visit\_page & Directly access a webpage by URL. \\ 
page\_up & Scroll up one page in the current view. \\ 
page\_down & Scroll down one page in the current view. \\ 
download\_file & Download and attempt to return the text of a file. \\ 
find\_on\_page\_ctrl\_f & Find and highlight text on a page using a search term. \\ 
find\_next & Move to the next occurrence of the search term on a page. \\ 
\bottomrule
\end{tabular}
\caption{Web Browser Tool Functions}
\label{tab:web_browser_functions}
\end{table}

Additionally, the web browser tool can convert the content of web pages into various formats for better processing, including plain text, HTML, and formats specific to sites like Wikipedia, YouTube, as well as document formats like DOCX, XLSX, PPTX, and multimedia formats such as WAV, MP3 (via ASR), and images (via OCR).

\paragraph{Computer Terminal}
The computer terminal tool serves as a code interpreter within the \emph{Sibyl} system. This tool allows the execution of Python code, facilitating dynamic computation and processing tasks which are crucial for complex problem solving and data manipulation within the AI framework.

These tools and their functionalities are important in enabling the \emph{Sibyl} agent to navigate, interpret, and interact with the digital world effectively, mirroring human-like web browsing and data processing capabilities.

\section{Ethics Statement}
By refining the ability to parse and process multifaceted information, \emph{Sibyl} helps in delivering more accurate and reliable outputs during complex reasoning in real-world scenarios. This progression helps improve the reliability of responses and reduce the occurrence of hallucinations in the outputs of these models. In addition, during the evaluation phase, we implemented rigorous monitoring of the reasoning processes to identify and prevent any potentially harmful actions that could be executed by the \emph{Sibyl} system. This precautionary measure is essential to ensure that while the system improves in autonomy, it remains within the bounds of ethical operation. Due to the versatility of general purpose LLM-based agents, we strongly recommend that users of \emph{Sibyl} also remain vigilant regarding the outputs produced, recognizing the potential impacts that these could have if misapplied. Users are urged to ensure that the deployment of \emph{Sibyl} is aligned with ethical standards and not used for malicious purposes.

\section{Prompts}
\label{sec:prompts}
\begin{longlisting}[Prompt Used in Tool Planner]
You are a helpful AI assistant.

I'll give you a question and a set of tools. Tell me which function you would use to solve the problem (or if you don't need any tool).

# Step History
{steps}

# Question
```text
{question}
```

# Tools

## Browser 
The functions of the browser will share the same session, that means the viewport will persist between calls
Every function will return the text of the current viewport after the action is performed. For long pages(longer than 1 viewport), you can use the page_up() and page_down() functions to scroll the viewport.
Since the page has been converted from HTML to Markdown, you cannot submit information using a form, nor can you enter information in any text boxes. If you want to use the form inside the page, try using the computer_terminal below to read the html content.
When the page is very long, content truncation may occur due to the limited display capacity of the viewport. You need to carefully consider whether additional page down is needed to ensure that you have obtained the complete information.
- informational_web_search(query: str) -> str:
    Perform an INFORMATIONAL web search query and return the search results.
- navigational_web_search(query: str) -> str:
    Perform a NAVIGATIONAL web search query and immediately navigate to the top result. Useful, for example, to navigate to a particular Wikipedia article or other known destination. Equivalent to Google's "I'm Feeling Lucky" button.
- visit_page(url: str) -> str:
    Visit a webpage at a given URL and return its text.
- page_up() -> str:
    Scroll the viewport UP one page-length in the current webpage and return the new viewport content.
- page_down() -> str:
    Scroll the viewport DOWN one page-length in the current webpage and return the new viewport content.
- download_file(url: str) -> str:
    Download a file at a given URL and, if possible, return its text. File types that will returned as text: .pdf, .docx, .xlsx, .pptx, .wav, .mp3, .jpg, .jpeg, .png(You can read the text content of the file with these extensions).
\end{longlisting}

\begin{longlisting}[Prompt Used in Tool Planner (continued)]
- find_on_page_ctrl_f(search_string: str) -> str:
    When the page is too long to be fully displayed in one viewport, you can use this function to scroll the viewport to the first occurrence of the search string. If the viewport has already displayed the entire page(Showing page 1 of 1.), there is no need to use this function. This is equivalent to Ctrl+F. This search string supports wildcards like '*'
- find_next() -> str:
    Scroll the viewport to the next occurrence of the search string.

## Computer Terminal
- computer_terminal(code: str) -> str
    You can use this function to run Python code. Use print() to output the result.

Based on the question and the step history, tell me which function you would use to solve the problem in next step. 
If you don't need any function or the question is very easy to answer, function "None" is also an option. 
Do not change the format and precision of the results (including rounding), as a dedicated person will handle the final formatting of the results.
Use JSON format to answer.
{format_instructions}
\end{longlisting}

\begin{longlisting}[Prompt Used for Improve Generated Code]
Your ultimate goal is to find the answer to the question below.
```text
{question}
```

# Step History
```text
{steps}
```

The next step is running the following code:
```python
{code}
```

Check this code and help me improve it.

Response in JSON format:
{format_instructions}
\end{longlisting}

\begin{longlisting}[Prompt Used in External Information Aquisition Channel]
Your ultimate goal is to find the answer to the question below.
```text
{question}
```

# Tools

## Browser
The functions of the browser will share the same session, that means the viewport will persist between calls
Every function will return the text of the current viewport after the action is performed. For long pages(longer than 1 viewport), you can use the page_up() and page_down() functions to scroll the viewport.
Since the page has been converted from HTML to Markdown, you cannot submit information using a form, nor can you enter information in any text boxes. If you want to use the form inside the page, try using the computer_terminal below to read the html content.
When the page is very long, content truncation may occur due to the limited display capacity of the viewport. You need to carefully consider whether additional page down is needed to ensure that you have obtained the complete information.
- informational_web_search(query: str) -> str:
    Perform an INFORMATIONAL web search query and return the search results.
- navigational_web_search(query: str) -> str:
    Perform a NAVIGATIONAL web search query and immediately navigate to the top result. Useful, for example, to navigate to a particular Wikipedia article or other known destination. Equivalent to Google's "I'm Feeling Lucky" button.
- visit_page(url: str) -> str:
    Visit a webpage at a given URL and return its text.
- page_up() -> str:
    Scroll the viewport UP one page-length in the current webpage and return the new viewport content.
- page_down() -> str:
    Scroll the viewport DOWN one page-length in the current webpage and return the new viewport content.
- download_file(url: str) -> str:
    Download a file at a given URL and, if possible, return its text. File types that will returned as text: .pdf, .docx, .xlsx, .pptx, .wav, .mp3, .jpg, .jpeg, .png(You can read the text content of the file with these extensions).
- find_on_page_ctrl_f(search_string: str) -> str:
    When the page is too long to be fully displayed in one viewport, you can use this function to scroll the viewport to the first occurrence of the search string. If the viewport has already displayed the entire page(Showing page 1 of 1.), there is no need to use this function. This is equivalent to Ctrl+F. This search string supports wildcards like '*'
- find_next() -> str:
    Scroll the viewport to the next occurrence of the search string.
\end{longlisting}

\begin{longlisting}[Prompt Used in External Information Aquisition Channel (continued)]
## Computer Terminal
- computer_terminal(code: str) -> str
    You can use this tool to run Python code. Use print() to output the result.

# Step History
```text
{steps}
```

# Current Step Tool Result
Tool: {tool}
Args: {args}
```
{tool_result}
```

# Instructions
1. Analyze the given tool result to extract relevant information directly contributing to answering the question.
2. Verify the information against the original question to ensure accuracy.
3. Record new facts only if they provide unique information not already found in the step history.
4. If the current tool result directly answers the question, record the answer and explain why no further steps are necessary.
5. If the current tool result is insufficient, plan a follow-up step to gather more data.
6. Choose the next tool and query that efficiently leads to the ultimate goal.
7. Minimize unnecessary steps by focusing on direct and efficient methods to gather required information.
8. Explain why you chose the next step and how it contributes to answering the question.
9. Do not change the format and precision of the results, as a dedicated person will handle the final formatting.
10. Your reply will be sent to the next agent for further action, so it is necessary to record all the information needed by the next agent in the plan (such as the complete URL of the link that needs to be clicked).

Response Format:
```text
Facts:
  1. Address: xxxx, Title: xxxx, Viewport position: xxxx
    xxxxx
  2. Address: xxxx, Title: xxxx, Viewport position: xxxx
    xxxxx
Explanation:
  xxxx 
Plan:
  xxxx 
```
\end{longlisting}

\begin{longlisting}[Prompt Used for Formatting Answers]
Format the following answer according to these rules:

1. **Numbers**:
   * If the answer contains a relevant number, return the number without commas, units, or punctuation.
   * If the number represents thousands, return the number in thousands.
   * Perform necessary unit conversions based on the context provided in the question. For example, convert picometers to Angstroms if the question implies this.
   * Retain the original precision of the number unless specific rounding instructions are given.
   * Numbers should be written as digits (e.g., 1000000 instead of "one million").

2. **Dates**:
   * If the answer contains a date, return it in the same format provided.

3. **Strings**:
   * Exclude articles and abbreviations.
   * Write digits in numeric form unless specified otherwise.
   
4. **Lists**:
   * If the answer is a comma-separated list, return it as a comma-separated list, applying the above rules for numbers and strings.

5. **Sentences**:
   * If the answer is a full sentence and the question expects a detailed explanation, preserve the sentence as is.
   * If the answer can be reduced to "Yes" or "No", do so.

Important:
1. Carefully interpret the question to determine the appropriate format for the answer, including any necessary unit conversions.
2. Return only the final formatted answer.
3. The final formatted answer should be as concise as possible, directly addressing the question without any additional explanation or restatement.
4. Exclude any additional details beyond the specific information requested.
5. If unable to solve the question, make a well-informed EDUCATED GUESS based on the information we have provided. Your EDUCATED GUESS should be a number OR as few words as possible OR a comma separated list of numbers and/or strings. DO NOT OUTPUT 'I don't know', 'Unable to determine', etc.

Here is the question:
{question}

Here is the answer to format:
{answer}

Formatted answer:
\end{longlisting}

\section{Limitations and Future Directions}

Despite the advancements demonstrated by the \emph{Sibyl} framework, there are inherent limitations that wait to addressed in the future.

\subsection{Limitations}
\paragraph{Lack of Vision Large Language Model Support} Currently, \emph{Sibyl} primarily operates on textual data and use Optical Character Recognition (OCR) to convert visual information into text modal, thus lacking integration with vision large language models, which restricts its ability to process and interpret visual content as humans do.

\paragraph{Browser Functionalities} While \emph{Sibyl} utilizes a browser tool, it is not yet equipped with a fully functional browser akin to those used by humans. This limitation affects the agent's ability to interact with web content in a more natural and efficient manner.

\paragraph{Learning Mechanisms} The present system does not incorporate learning mechanisms to adapt and improve from real-world interactions on-the-fly. This may restrict its ability to evolve based on new data or scenarios it encounters.

\subsection{Future Directions}

\paragraph{Integrating Vision Large Language Models} Future versions of \emph{Sibyl} will incorporate vision large language models to allow the system to handle multimedia content effectively, broadening its applicability across various domains where visual data plays a crucial role.

\paragraph{Enhancing Browser Capabilities} There is a pressing need to optimize the browser tool to provide full functionality, mirroring the capabilities available to human users, thereby improving the agent's interaction with web interfaces.

\paragraph{Designing Adaptive Learning Mechanisms} In the future version of \emph{Sibyl}, we plan to introduce adaptive learning mechanisms will enable the system to learn from its interactions and experiences, thereby progressively improving its problem-solving strategies and effectiveness.

\paragraph{Developing LLMs tailored for Agents} Future work will also focus on developing LLMs that are specifically optimized for general-purpose AI agents, aiming to enhance their efficiency and effectiveness in complex reasoning tasks. This includes gathering and utilizing data specific to long-distance reasoning processes in real-world scenarios and improving the system’s ability to build and reuse its tools autonomously.

\end{document}